\documentclass[conference]{IEEEtran}
\usepackage{authblk} 

\usepackage[utf8]{inputenc}

\usepackage[backend=biber,style=numeric]{biblatex}
\usepackage{amsmath}
\usepackage{booktabs}
\usepackage{graphicx}
\usepackage[main=british]{babel}
\usepackage[babel=true]{csquotes}
\usepackage{mwe}

\usepackage{times}
\usepackage{latexsym}
\usepackage{verbatim}
\usepackage[utf8]{inputenc}
\graphicspath{ {./figures/} }
\usepackage{subcaption}
\usepackage[T1]{fontenc}
\usepackage[scaled]{beramono}
\usepackage{multirow}
\usepackage{authblk}
\usepackage[amsmath,thref,hyperref]{ntheorem}
\usepackage{bookmark}
\usepackage{hyperref}
\usepackage{silence}
\begin{document}
\title{Subgraph2vec: A random walk-based algorithm for embedding knowledge graphs}

\author{Elika Bozorgi$^{1,*}$, Saber Soleimani$^1$, Sakher Khalil Alqaiidi$^1$, Hamid Reza Arabnia$^1$, Krzysztof Kochut$^1$}

\affil{
$^{1}$School of Computing, The University of Georgia, Athens, GA, USA \\
\href{}{ {\{elika.bozorgi, saber.s, sakher.a, hra, kkochut\}}@uga.edu}
}

\maketitle
\begin{abstract}

Graph is an important data representation which occurs naturally in the real world applications \cite{goyal2018graph}. Therefore, analyzing graphs provides users with better insights in different areas such as anomaly detection \cite{ma2021comprehensive}, decision making \cite{fan2023graph}, clustering \cite{tsitsulin2023graph}, classification \cite{wang2021mixup} and etc. However, most of these methods require high levels of computational time and space. We can use other ways like embedding to reduce these costs. Knowledge graph (KG) embedding is a technique that aims to achieve the vector representation of a KG. It represents entities and relations of a KG in a low-dimensional space while maintaining the semantic meanings of them. There are different methods for embedding graphs including random walk-based methods such as node2vec, metapath2vec and regpattern2vec. However, most of these methods bias the walks based on a rigid pattern usually hard-coded in the algorithm. In this work, we introduce \textit{subgraph2vec} for embedding KGs where walks are run inside a user-defined subgraph. We use this embedding for link prediction and prove our method has better performance in most cases in comparison with the previous ones. 
\hfill \break

Keywords: Representation Learning, Information Engineering, Link Prediction, Deep Learning, Graph Embedding
\end{abstract}

\section{Introduction}

Knowledge graphs play a crucial role in organizing, understanding, and leveraging information in various domains. Hence, they become increasingly popular in different areas due to their valuable features.
For example, they are widely used in recommendation systems in E-commerce to model the relationships between users and items. Next, they enable personalized recommendations by leveraging knowledge about user interactions and item characteristics \cite{huang2004graph}. As another example, they are widely used in healthcare to support practitioners for disease diagnosis by knowledge discovery from patients’ personal health repositories \cite{tao2020mining}.   
In addition, they provide a structured way of connecting entities with relationships, which creates a network to organize and represent information. Thus, it is easier to discover relevant information either manually, by navigating through nodes to discover visions, or by using machine learning and AI to make predictions and generate insights into data. In addition, Knowledge graphs facilitate querying and discovery of complex patterns within data \cite{verborgh2016triple} and also capture semantic connections which leads to more accurate queries. 
 
Although knowledge graphs offer many benefits, they have several issues. For instance, Knowledge graphs are often incomplete \cite{chen2022fuzzy}, as they are frequently populated using various external resources which are often incomplete as well. Also, these outside resources have different structure and format which makes integrating data more difficult. In addition, knowledge graphs usually grow in size and complexity which makes them inefficient and therefore special infrastructure and algorithms are needed to make them more scalable \cite{polleres2023does}. This complexity also makes interpreting patterns and drawing insights more challenging and time-consuming. Moreover, building and maintaining knowledge graphs often need domain expertise to ensure the accuracy of the represented data.

A reasonable number of such issues can be addressed using Artificial Intelligence (AI). In fact, different methods of AI can be used for resolving the previous mentioned problems with knowledge graphs as well as: knowledge graph completion \cite{chen2020knowledge}, node representation learning \cite{zhang2021labeling}, semantic search \cite{reinanda2020knowledge}, question answering \cite{saxena2020improving}, anomaly detection \cite{ma2022deep}, quality assessment \cite{huaman2022steps}, etc. However, applying AI algorithms on different types of Knowledge graphs requires a significant number of adjustments due to the high number of the dimensions of the input data. In fact, in many cases, these methods cannot be applied on Knowledge graphs directly. Therefore, it is desired to reduce the number of the dimensions of the input data or knowledge graph by embedding it with different AI methods. Knowledge graph embedding is the act of translating the high-dimensional data to a low-dimensional space, while trying to maintain the semantic meanings of the KG elements.
The embedding of each element in the dataset is a unique vector-representation of that element. The resulted embedding of the Knowledge graph can be used for different purposes, such as link prediction, entity classification, semantic search, and others.
There exist different types of embedding methods for Knowledge graphs based on supervision, which include \textit{supervised} methods such as Graph Convolutional Networks (GCNs) \cite{yao2019graph}, \textit{Unsupervised} methods like TransE \cite{bordes2013translating} and DeepWalk and \textit{Hybrid} ones such as GraphSAGE \cite{hamilton2017inductive}.
In this paper, we introduce an unsupervised algorithm based on random walks for embedding Knowledge graphs. There are previous random walk-based methods for embedding knowledge graphs, such as node2vec, metapath2vec and regpattern2vec. However, these methods come with challenges. For example, in node2vec the random walks are biased to highly visible types of nodes, the ones with
a dominant number of paths. On the other hand, in metapath2vec and regpattern2vec the walks are biased by a series of relationships (or node types) or a fixed regular pattern of relationships (or node types), respectively. In our method, the user enters an arbitrary pattern which defines a schema subgraph in the actual knowledge graph and the walk is done within this subgraph. In the next section, we will compare our method with the previous related ones.

\section{Preliminaries}
In this section, we will explain some primitive concepts which are fundamental to the understanding of our method and the previous ones, starting with the explanation of the Knowledge graphs (or heterogeneous networks).
\hfill \break

\textbf{Knowledge graph}: A knowledge graph is a data set that represents real-world facts and semantic relationships in the form of triplets, where the triplets are represented as a graph with edges as relations and nodes as entities \cite{bordes2011learning}. Mathematically, consider $G = (V,E)$ where $G$ represents the knowledge graph and $V$ are the nodes or entities and $E$ represent the relations.
\hfill \break
\textbf{Walk:} A walk is a finite sequence of edges which join a sequence of vertices. In $G = (V,E)$ where $G$ is a knowledge graph and $V$ represents the nodes and $E$ represents the edges, a finite walk is a sequence of edges $(e_{1}, e_{2}, ..., e_{n-1})$ for which there is a sequence of vertices $(v_{1}, v_{2}, ..., v_{n})$ such that $\phi(e_{i}) = (v_{i}, v_{i+1})$ for $i = 1, 2, ..., n-1$.  $(v_{1}, v_{2}, ..., v_{n})$ is the vertex sequence of the walk. The walk is closed if $v_{1} = v_{n}$, and it is open otherwise \cite{enwiki:1191420024}. 
\hfill \break
\textbf{Path:} A path is a walk on a graph where the vertices are repeated.
\hfill \break
\textbf{Subgraph:}\textit{}
Graph ${S = (V_S, E_S)}$ is considered as a subgraph of ${G = (V_G, E_G)}$ if and only if its vertex set $(V_S)$ and edge set $(E_S)$ are subsets of those of G. In other words: $V_S \in V_G$ and $E_S \in E_G$. 
\hfill \break
\textbf{Schema:}\textit{}
Given a knowledge graph $G$, an edge with a relation type $R$ connects source nodes of
type $S$ and target nodes of type $T$ defines a meta edge
$S \xrightarrow R \hbox{$T$} $ . A schema graph (aka known as meta-template) for $G$ is a set of all such meta edges. In fact, a schema
graph is a directed graph defined over node types $T$, with edges from $R$, denoted as $G_S = (T, R)$ \cite{sipser2012introduction}.
\hfill \break
\textbf{Heterogeneous Network \cite{yang2018meta}:}\textit{} or HIN, is a graph denoted as $ G = (V,E,T)$, where each $v \in V$ and $e \in E$ has a mapping function $\Phi (V) = V \rightarrow T_v$ and $\phi (E) = E \rightarrow T_e$ and $T_v$  and $T_e$ denote sets of node and relation types where $|T_v|+|T_e| > 2$. In simple words, in these networks, nodes and edges can belong to different types, and the connections between nodes can have various semantic meanings.

\section{Related works}

\textbf{DeepWalk} \cite{perozzi2014deepwalk}, is a graph embedding method which aims to learn continuous representations of nodes in a graph. For each walk, it begins by generating the walks from a random starting node and moves to the next random node on the graph. Each random walk sequence is treated as a sentence, in which the nodes are considered as the words of the sentence. Next, the word2vec \cite{church2017word2vec} model from NLP is applied to learn embeddings for the nodes. The resulting sentences are embedded by giving them as an input to a Skip-gram model, a variant of the one used in the word2Vec model. After the skip-gram is trained, it predicts the context nodes based on its input and the output is the embedding of the nodes.
\hfill \break
\textbf{Node2vec} \cite{grover2016node2vec}, is a technique for learning node embeddings in a graph. In fact, it is an extension of the word2vec model which is a method for word embedding in textual data. It learns embeddings of the nodes in a graph by using a neighborhood sampling strategy which captures both local and global structures. Node2vec generates random walks which are balanced between breadth-First Search (BFS) and depth-First search (DFS) and applies it on the input data. After generating random walks, node2vec learns embeddings using the Skip-gram model. The Skip-gram model predicts the context (neighboring nodes) of a target node based on its embedding. Node2vec optimizes the obtained embeddings by maximizing the likelihood of observing neighboring nodes within the context window of each target node.
\hfill \break
\textbf{Metapath2vec} \cite{dong2017metapath2vec}, is a representation learning method designed specifically for HINs to learn embeddings of the nodes and captures both semantic and structure of the network.
In this method, meta-path guided random walks are used to make a sentence of the nodes. 
The meta-paths are created by domain experts based on nodes according to the dataset. For example, on a DBLP computer science bibliographic dataset \cite{ley2006maintaining}, the created meta-paths are : APA, APVPA, OAPVPAO, where A represents the author, P the paper, O the organization and V the venue. Consider APA as an example. The first node to choose must be of type A, the second node must be of type P and the third node must be of type A.
This means that at each step the next node to visit is chosen according to a pre-defined meta-path (APA in this example), ensuring that the walk follows a meaningful path in the network. After generating a large number of meta-path based random walks, these sequences of nodes are used as training data for the Skip-gram model. Finally, these embeddings are aggregated to obtain comprehensive representations for each node in the network beside capture the diverse relationships and semantic meanings associated with each node. 
\hfill \break
\textbf{Regpattern2vec} \cite{keshavarzi2021regpattern2vec}, is a method for embedding KGs which samples a large knowledge graph to learn node embedding while capturing the semantic relationships among the nodes of the graph.  
In this method, the walk is biased by a fixed pattern \textit{H[\textasciicircum T] + HT} which is based on edges. The walk starts at a random node from a set of given nodes known as source nodes $(S)$ on the knowledge graph. After choosing the source node, the walk chooses an edge of type $H$ and moves to a randomly chosen neighbor creating a path at each time. Next, the walk chooses a random edge of type \textasciicircum \textit{T} and moves to the next random node. The walk follows the pattern \textit{H[\textasciicircum T] + HT}  and continues the walk according to the a parameter called \textit{walk length}.  We can control the number of the walks in each path by \textit{walk length} and when the number of the walks reaches to \textit{walk length}, the walk is complete. 
Once the first walk is complete, we can start another walk by choosing a random node from $S$. The number of the times we start a new walk is based on a parameter called \textit{number of walk}.
Choosing a node from the set of source nodes each time as the starting node and creating random walks will result in several paths of random walks. To obtain the embedding of these nodes, the resulted paths are fed into a modified version of a skip-gram model. This method of embedding is an unsupervised method and it can be applied on any given raw text corpus or document.
In regpattern2vec, the algorithm runs the walks on a rigid regular expression H[\textasciicircum T] + HT which is defined by domain experts and is hard-coded in the algorithm and cannot be changed.

\hfill \break
\textbf{Contribution:} In the previously mentioned methods, the algorithms are either based on a pre-defined sequence of node/edge types or a pattern designed by domain experts or is biased toward specific nodes by experts. It means that they are not generic and the user does not have any role in guiding the walks. In our algorithm, however, we define a method in which the algorithm runs on any arbitrary random walk path inside a user-defined schema subgraph based on edges.
The user enters a schema subgraph by entering integers where each integer represents an edge in the knowledge graph. This schema subgraph defines the subgraph. After the subgraph is defined, we choose the first random node as the source node inside this subgraph and continue the walk based on the \textit{walk length} parameter and move to any random nodes via any random edges. The walks are valid only if they are within the subgraph and invalid otherwise.
The advantage of using a subgraph is that it is more permissive; since we can run the walks totally randomly inside the user-defined subgraph rather than having biased walks based on a rigid pattern like the previous mentioned methods. The walks are valid as long as they are within the subgraph and invalid otherwise. 
 
\section{Methodology}
\hfill \break
Our algorithm runs with any user-given schema subgraph ($s$) based on the edges. The schema subgraph is entered in the form of integers where each integer denotes an edge of the KG. This schema subgraph which defines the subgraph ($S$) is actually a part of the original graph ($G$). 
Let's assume the user enters a schema subgraph such as $s$ = '$x_1, x_2, x_3$' where $x_{i=1,2,3}$ is an integer representing an edge based on the dataset.  

This schema subgraph defines subgraph $S = (V', E') \in G = (V, E)$ where $G$ is the primary knowledge graph. The algorithm chooses the first node inside $S$ randomly and from there, uses the below equation to calculate the probability of each neighbor edge based on its type ($t$). 
To calculate the probability of moving to the next edge based on subgraph $S$, we use this equation:

\begin{equation*}
\sum^{l}_{i=1+1} P(r^{i+1}|r^{i}, S) = 
\end{equation*}

\begin{equation*}
\begin{aligned}
\begin{cases}
\frac{1}{r_{t_i}} \times \frac{1}{\sum^{n}_{i=1} t_i} \; \; \; \; \ \ \ \ \ \ \ \ \ \ \ \ \ \ \ \ \ \ \ \ \ \ \ \ \ \ \ \ \ \ \ \ \ \left(r \in S \right) \\ \\ 

0  \ \ \ \ \ \ \ \ \ \ \ \ \ \ \ \ \ \ \ \ \ \ \ \ \ \ \ \ \ \ \ \ \ \ \ \ \ \ \left(r^i,\ r^i,\ r^{i+1}\right)\notin \ G'
\end{cases}
\end{aligned}
\end{equation*}
\textbf{}
\hfill \break

where $t_i$ denotes each type of edges connected to the current node and  $r_{t_i}$ denotes the number of the edges of each type. We choose the type of the next edge based on its probability. An important thing to consider is that we have a hierarchy of edges in our knowledge graph; which means that a type of edge might have different sub-types. With that being said, all the sub-types of a specific type should be considered as the same type. 




In addition, we have a parameter called \textit{number of walks} in our code, which defines the number of the walks that should be walked in each path. The default number for it is set as 40, which the user can adjust on their own interest in the code.



    


\begin{figure*}
       \centering

\begin{subfigure}[b]{0.24\textwidth}
\includegraphics[width=\textwidth]{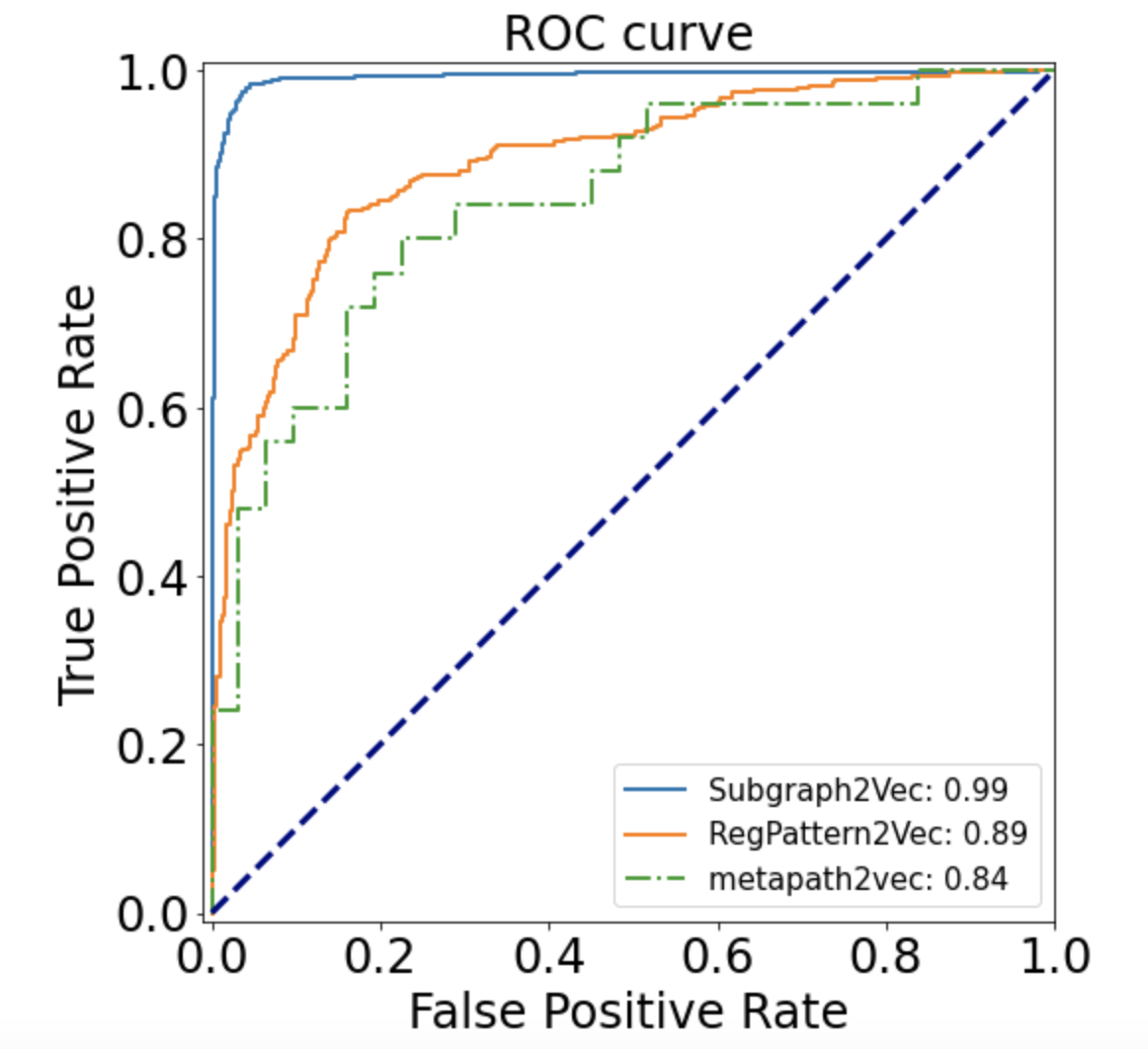}
\caption{ROC on YAGO Dataset, predicted link:is citizen of}
\end{subfigure}
\begin{subfigure}[b]{0.24\textwidth}
\includegraphics[width=\textwidth]{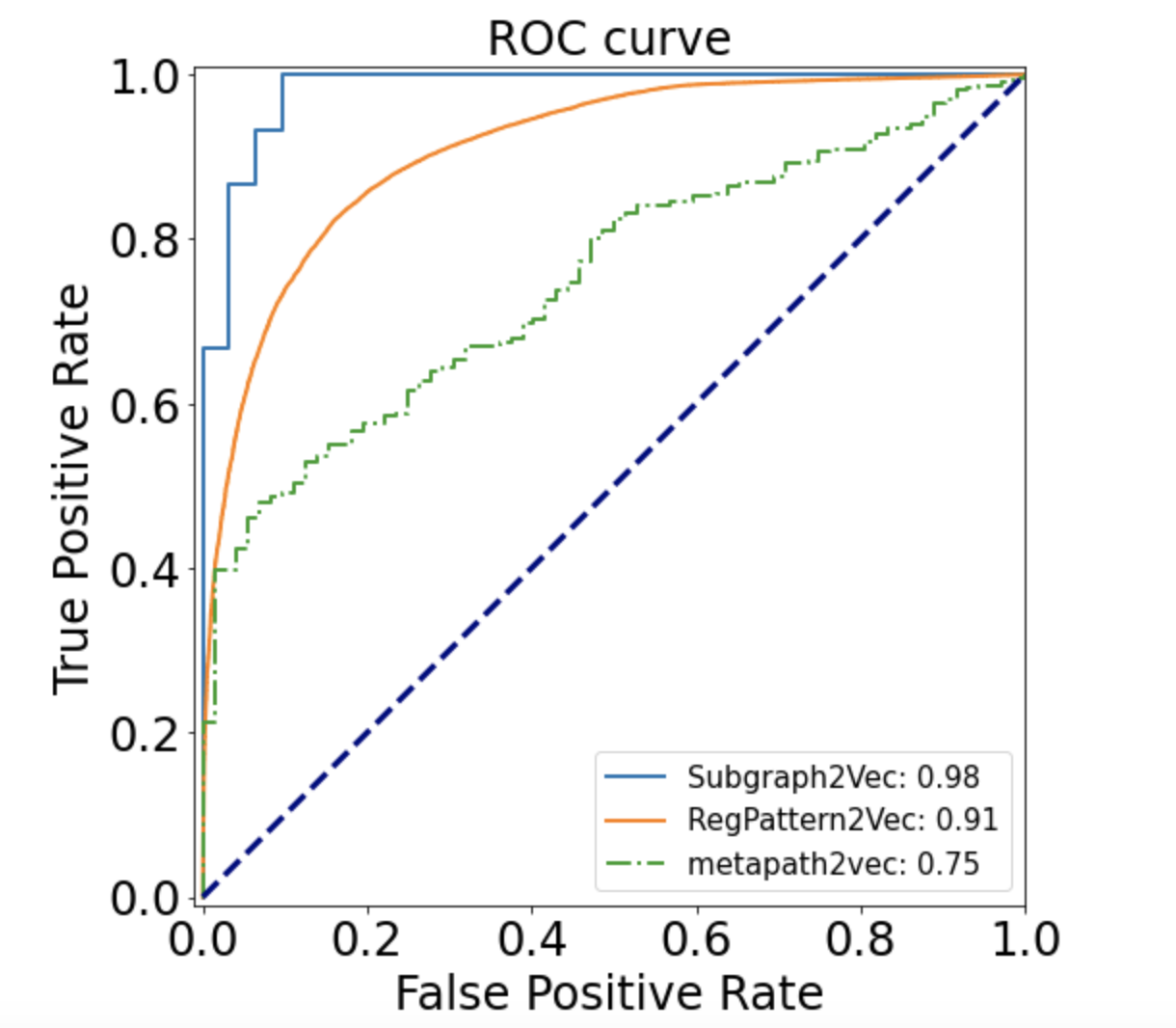}
\caption{ROC on YAGO Dataset, predicted link:Located In}
\end{subfigure}
\begin{subfigure}[b]{0.24\textwidth}
\includegraphics[width=\textwidth]{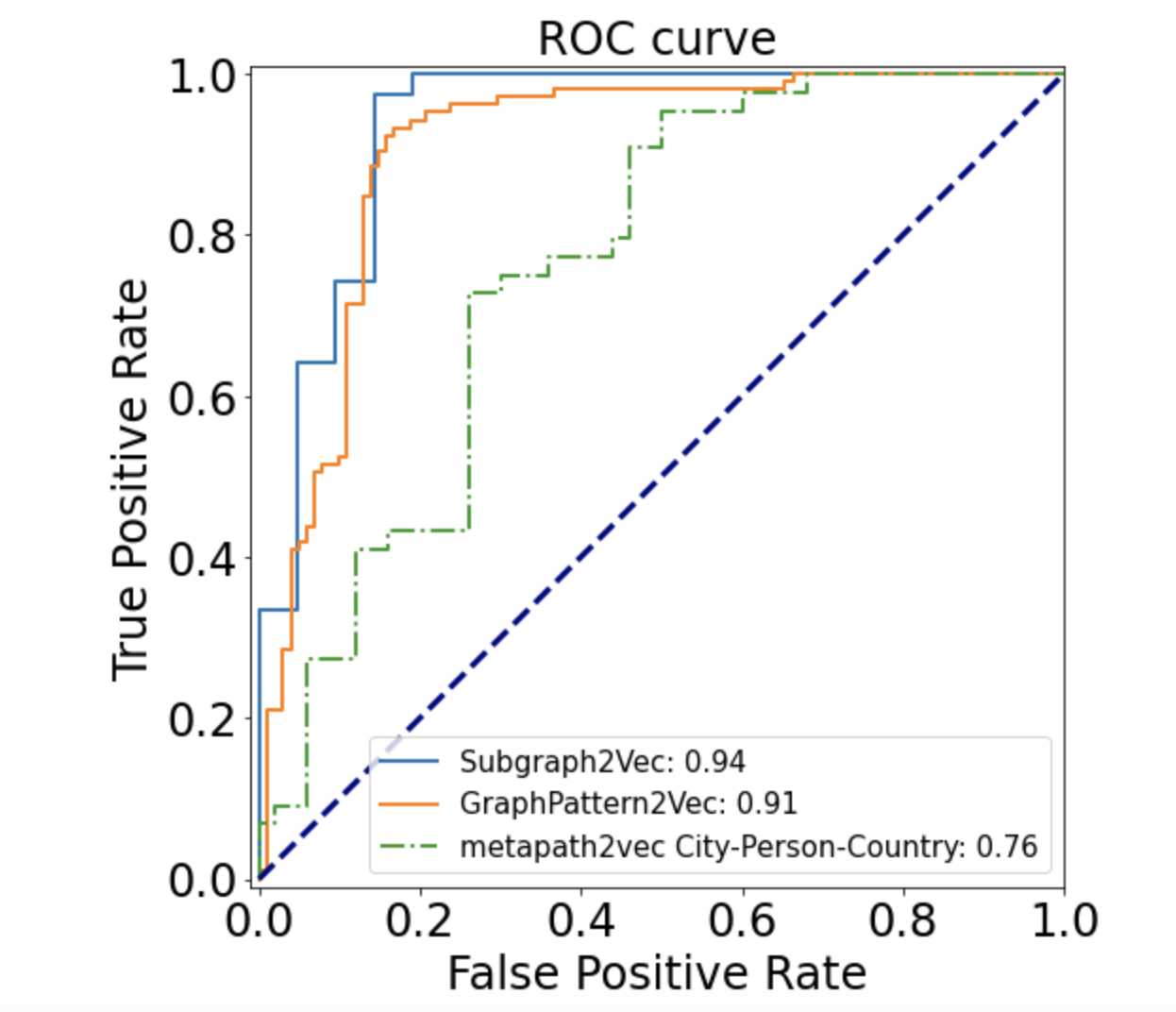}
\caption{ROC on YAGO Dataset, predicted link:Leader of}
\end{subfigure}
\newline
\begin{subfigure}[b]{0.24\textwidth}
\includegraphics[width=\textwidth]{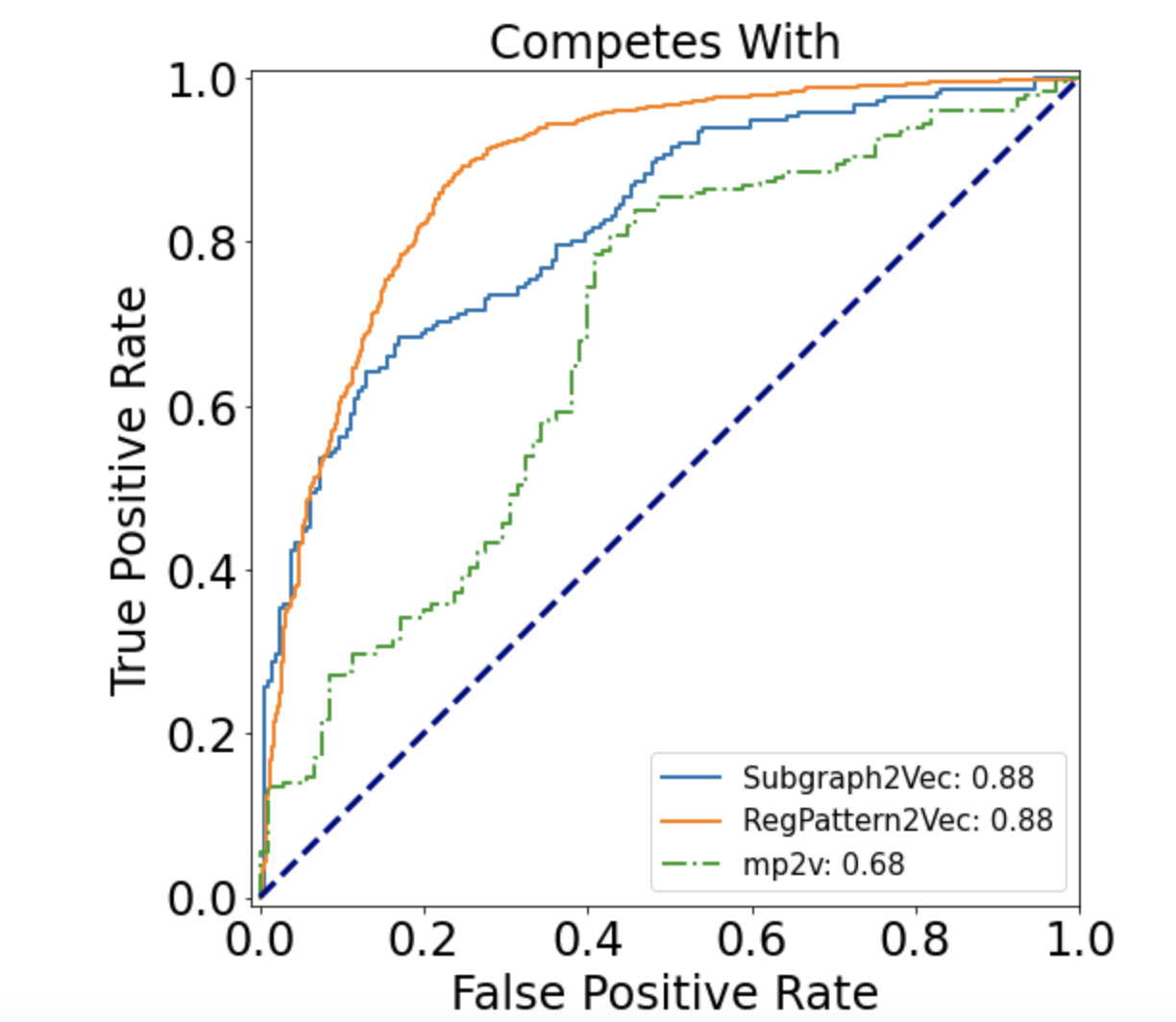}
\caption{ROC on NELL Dataset, predicted link:competes with}
\end{subfigure}
\begin{subfigure}[b]{0.24\textwidth}
\includegraphics[width=\textwidth]{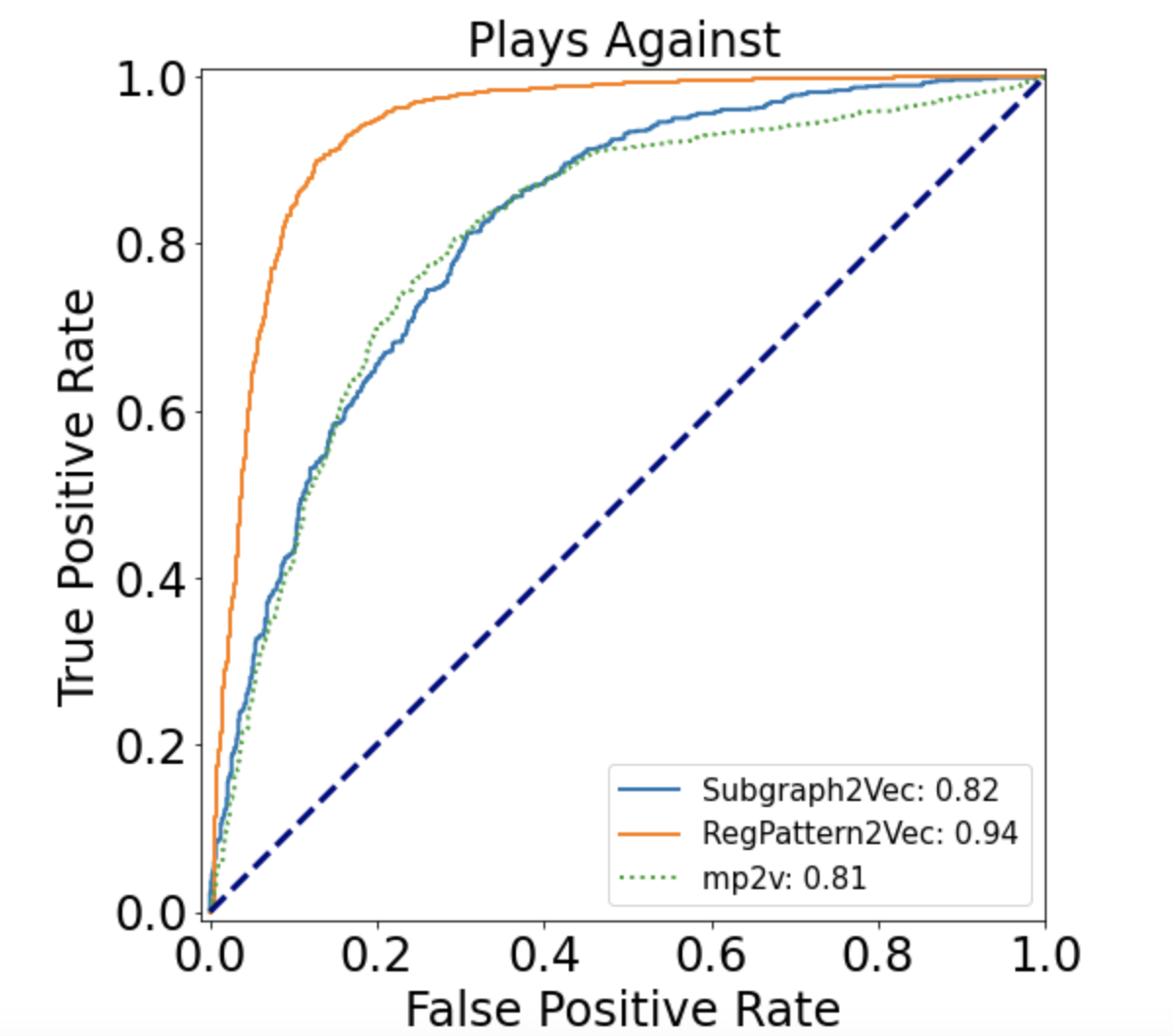}
\caption{ROC on NELL Dataset, predicted link:plays against}
\end{subfigure}
    \caption{Comparing ROC of different links of Subgraph2vec, Regpattern2vec and Metapath2vec on NELL and YAGO.}
    \end{figure*}

\subsection{Walk-based Embedding}
After the user has entered the schema subgraph, the subgraph ($S$) is defined and we are able to conduct the walk. We start the walk by choosing the first node randomly within the sub-graph. Then we will choose an edge connected to this node randomly and if it is inside the schema-graph, the chosen edge is valid which means we choose the next random edge. Otherwise, we will delete that edge type from the neighbor of the node and choose another random edge and check its validation. We have decided to conduct the walk based on 40 steps (walk\_length $=$ 40). 
We should repeat this scenario for 40 walks (number\_of\_walks $=$ 40) and these walks are written to a file. Both walk-length and number\_of\_walks can be modified by the user.
After traversing all the walks and writing them on the file, we obtain the walk file which contains all the walks and their steps and then we should embed these walks. Embedding, in simple words is translating high-dimensional vectors to a low-dimensional space. To embed the walks of the walk file, we can consider each walk as a sentence of the words in which each node is considered as one word. Therefore, we can get help from the word2vec model in which a neural network learns node embedding from a corpus. 

In this paper, we are using a modified version of the skip gram model which captures the similarity of the walks based on their types. Skip gram is from the word2vec family and all of the word2vec models are consisting of two-layer neural networks used for word embedding. In a general sense, in the skip gram architecture, the model uses the current word (input) to predict the surrounding window -usually of size 5 to 10- of the context words (output). In fact, a skip gram is trying to find a semantic similarity between the words in a context by learning a meaningful representation of each word (embedding) in the document.  After feeding the walk file as an input to the skip gram model and getting the embedding file, we are able to use it. The embedding file can be used for various tasks such as link prediction, node classification, community detection and etc. In this work, we use the embedding file for link prediction.

        
       
        
        
        
        
    

\subsection{Application} 
In this paper, we decided to move forward with the link prediction task.
For conducting link prediction, we need to train our model first. We do it by using the vector representation of the current edges in the graph which is considered as the positive example. Therefore, for negative examples, we can consider combining pairs of edges in the graph that are not connected. Both of the positive and negative examples are used to train the classification model. We use Logistic Regression as the classifier, which can be used for link prediction as well. 



\section{Experiments}
\hfill \break
In this section, we will evaluate the conductance of the subgraph2vec method by running it on different datasets.
\subsection {Dataset} We use two different datasets to evaluate our model.

(i) The first dataset is YAGO39K \cite{lv2018differentiating}, which includes data from Wikipedia, WordNet and GeoNames and is a subset of the YAGO knowledge base \cite{chirita200707}. It contains 123,182 unique entities and 1,084,040 unique edges with 37 different relation types. 
(ii) The second dataset used was NELL, which is built from the Web via an intelligent agent and contains 49,869 unique nodes 296,013 edges and 827 relation types.

\begin{table}
   \caption{Statistics of split data (MST method) based on the relation to be predicted \cite{keshavarzi2021regpattern2vec} } 
   \small
   \centering
   \begin{tabular}{lccr}
   \toprule\toprule
   \textbf{Dataset } & \textbf{Relations} & \textbf{Train est} & \textbf{Test set} \\ 
   \midrule
   \multirow{2}{*}{NELL} & CompetesWith & $9,154$  & $1,070$ \\
                    & PlaysAgainst & $2,945$   & $2,225$ \\
   \midrule
   \multirow{2}{*}{YAGO} & isLocatedIn & $44,542$  &$44,541$\\
                                    & isCitizenOf & $3,128$   & $342$ \\
                                    & isLeaderOf  & $855$ & $106$ \\
   
   \bottomrule
   \end{tabular}
    
\end{table}

\subsection {Experimental Setup}
To apply random walks in Subgraph2vec, we set the number\textunderscore of\textunderscore walks = 40 and maximum walk\textunderscore length = 40.
Also, the logistic regression parameters are constant for all the datasets.

For both of our datasets, we split the dataset in order to train and test for any relation we want to predict in either of our datasets. Note that in our model, it is necessary to have the pair of the nodes we want to do prediction on in both the training and the test dataset. However, the relation between that pair is different in each of the train and test dataset. To achieve this goal, we apply minimum spanning tree method to take the minimum possible nodes from the graph to prepare the test dataset.

We have split the dataset for each of the relations we want to predict individually. Table 1 illustrates the number of the rows for each dataset (split with MST) based on the relation to be predicted.
\subsection{Link Prediction}

To explain the link prediction, we will give a brief explanation for each dataset.
For the YAGO dataset, we decided to predict these relations: 'isLocatedIn', 'isCitizenOf', 'isLeaderOf'. Here is $S$ defined for these relations: 

For 'isLeaderOf', $S$ consists of these edges: 'PlayisIn', 'isLeaderOf' and 'isLocatedIn'. For 'isCitizenOf', $S$ consists of these edges: 'isCitizenOf', 'isLocatedIn' and 'isLocatedIn'. For 'isLeaderOf', $S$ consists of these edges: 'isLeaderOf', 'isLocatedIn' and 'wasBornIn'. 

For the NELL dataset, we chose two relations of interest: 'competesWith' and 'playsAgainst'. 
For 'competeswith', $S$ consists of :'Competeswith', 'hasofficeincity' and 'cityhascompanyoffice'.
For 'playsAgainst', $S$ consists of :'teamplaysagainstteam' or in short 'playsagainst' which is the link to be predicted, 'teamplaysinleague' and 'sportsgameteam'.

Figure 1 shows the prediction results on the test datasets.
Here, we are comparing the results of our method to the results we obtained from running regpattern2vec and metapath2vec algorithms using Nell and YAGO data sets. Our results imply that our method outperforms the 
regpattern2vec and metapath2vec methods in most cases. That is due to being capable of choosing the nodes/edges randomly within the subgraph rather than choosing them based on a regular expression. Figure 1 illustrates the ROC curve from each of the algorithms.

\section{Conclusion and future work}
\hfill \break
In this paper, we present Subgraph2vec, a random walk-based method in which a subgraph is used for limiting random walks on a knowledge graph in a generic fashion.
There are different random walk-based methods for embedding knowledge graphs such as node2vec, metapath2vec and regpattern2vec which were discussed earlier. In the previous methods, random walks are biased based on different algorithms such as BFS and DFS in node2vec or fixed patterns in metapath2vec and regpattern2vec. However, our goal is to implement an algorithm which runs the random walks inside on a user-given subgraph; which makes the embedding algorithm more broad. 
The generated walk file is embedded using skipgram and the resulted embedding files can be used for various purposes such as node classification, link prediction, community detection and etc. In this work, we decided to use it for link prediction.
Our results on NELL and YAGO datasets prove our method outperforms other methods such as regpattern2vec and metapath2vec in most cases.
In the future work, we will use the obtained embedding file for the mentioned tasks and also in another work we will add weights to the edges to make our model more customizable.
\hfill \break

\printbibliography

\end{document}